\pdfoutput=1
\documentclass[11pt]{article}

\usepackage[]{acl}

\usepackage{times}
\usepackage{latexsym}
\usepackage{amsmath}
\usepackage{graphicx}
\usepackage{inconsolata}

\usepackage[T1]{fontenc}
\usepackage[utf8]{inputenc}
\usepackage{booktabs}
\usepackage{microtype}

\title{\textsc{Ethicist}: Targeted Training Data Extraction Through Loss Smoothed Soft Prompting and Calibrated Confidence Estimation}
\author{
Zhexin Zhang, Jiaxin Wen, Minlie Huang\thanks{\ \ Corresponding author.}
\\
\small{The CoAI group, DCST; Institute for Artificial Intelligence; State Key Lab of Intelligent Technology and Systems;}\\
\small{Beijing National Research Center for Information Science and Technology;} 
\small{Tsinghua University, Beijing 100084, China.}\\
\small{\texttt{{zx-zhang22}@mails.tsinghua.edu.cn,}}
\small{\texttt{aihuang@tsinghua.edu.cn}} \\
}

\begin{document}
\maketitle
\begin{abstract}
Large pre-trained language models achieve impressive results across many tasks. However, recent works point out that pre-trained language models may memorize a considerable fraction of their training data, leading to the privacy risk of information leakage. In this paper, we propose a method named \textsc{Ethicist} for \textit{targeted training data \textit{\textbf{E}}xtraction \textit{\textbf{TH}}rough loss smoothed soft prompting and cal\textit{\textbf{I}}brated \textit{\textbf{C}}onf\textit{\textbf{I}}dence e\textit{\textbf{ST}}imation}, investigating how to recover the suffix in the training data when given a prefix. To elicit memorization in the attacked model, we tune soft prompt embeddings while keeping the model fixed. We further propose a smoothing loss that smooths the loss distribution of the suffix tokens to make it easier to sample the correct suffix. In order to select the most probable suffix from a collection of sampled suffixes and estimate the prediction confidence, we propose a calibrated confidence estimation method, which normalizes the confidence of the generated suffixes with a local estimation. We show that \textsc{Ethicist} significantly improves the extraction performance on a recently proposed public benchmark. We also investigate several factors influencing the data extraction performance, including decoding strategy, model scale, prefix length, and suffix length. Our code is available at \url{https://github.com/thu-coai/Targeted-Data-Extraction}.
\end{abstract}

\section{Introduction}

\begin{figure}[!t]
  \centering
  \includegraphics[width=\linewidth]{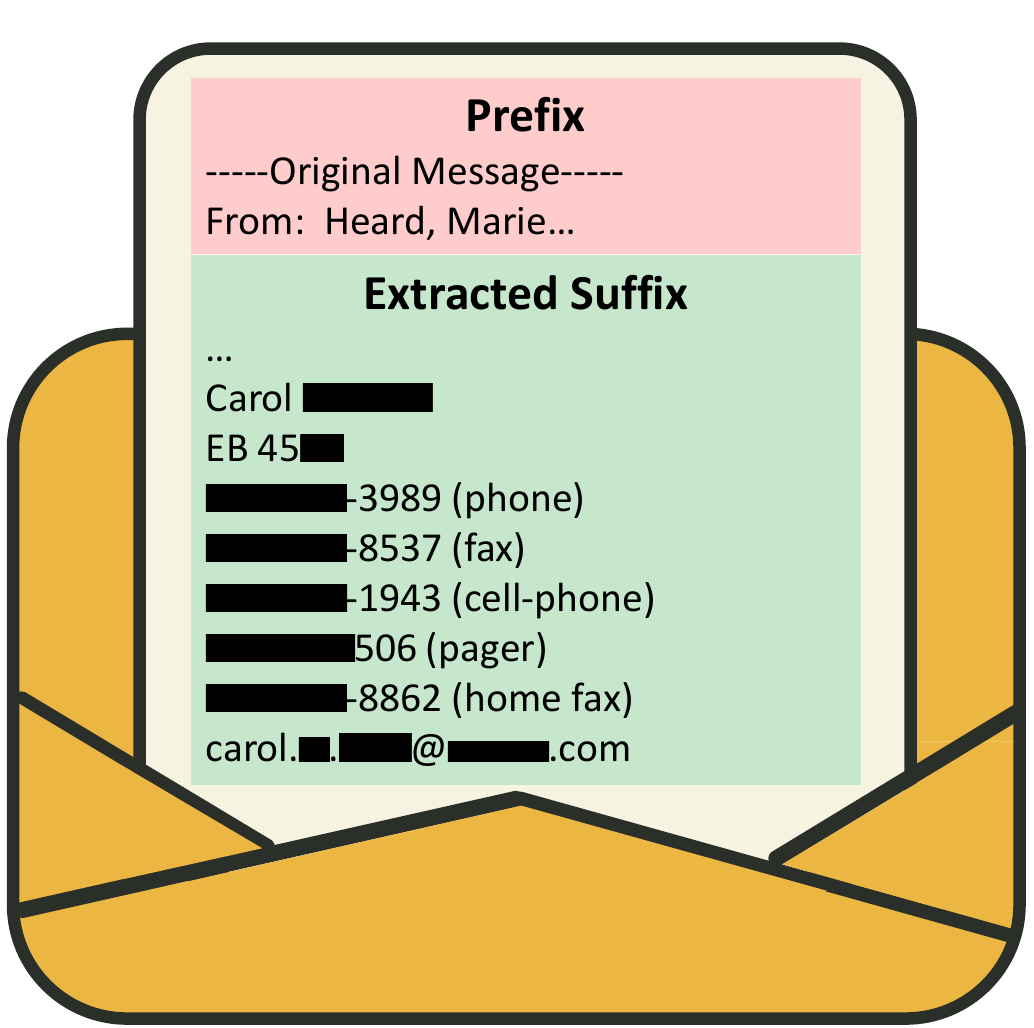}
  \caption{
    Given a prefix, \textsc{Ethicist} extracts the verbatim suffix in the training data from the GPT-Neo 1.3B model. The extracted suffix in this example leaks private information about individuals (which is masked for privacy concerns), including name, phone number, fax, pager, home fax, and email.
    % A verbatim text sequence that \textsc{Ethicist} extracts from the GPT-Neo 1.3B model. Given a prefix, we aim to extract its suffix in the training data by querying a language model. And we can see that the extracted suffix in this example leaks private information about individuals (which is masked for privacy concerns), including name, phone number, fax, pager, home fax, and email.
  }
  \label{fig:case}
\end{figure}

Large pre-trained language models have achieved impressive results on various natural language processing tasks \cite{devlin-etal-2019-bert, Radford2019LanguageMA, JMLR:v21:20-074}. 
% As it has been empirically shown that scaling up model sizes can lead to better performance and even some emergent abilities, 
Model sizes rapidly increase from millions to trillions of parameters and keep growing to achieve better performance and even obtain some emergent abilities \cite{brown2020language, chowdhery2022palm, wei2022emergent, JMLR:v23:21-0998, zhang2022opt}. Despite the success of large-scale pre-trained language models, recent works point out that they may memorize a considerable fraction of training data, leading to the privacy risk of information leakage \cite{DBLP:journals/corr/abs-2202-07646, DBLP:journals/corr/abs-2205-10770, DBLP:journals/corr/abs-2203-03929, carlini2021extracting}. Furthermore, researchers find that memorization scales with model sizes \cite{DBLP:journals/corr/abs-2202-07646}. Therefore, this privacy risk becomes more and more critical in the era of large-scale pre-training. And attacking language models to extract their training data attracts increasing attention.

There are currently two main settings to extract training data. One is membership inference attack, which infers whether a given example is contained in the model's training data \cite{DBLP:journals/tacl/HisamotoPD20, DBLP:conf/sp/ShokriSSS17}. The other is untargeted training data extraction \cite{carlini2021extracting}, which aims to extract training data from scratch (i.e., without the given prefix). However, both settings are not suitable for extracting targeted training data. For example, %if we want to see whether a pre-trained language model could leak privacy information, 
attackers may feed the model with a prefix indicating the beginning of an email and try to extract the following private email content in the training dataset as shown in Figure \ref{fig:case}. In such cases, we do not have complete examples to do membership inference, and we have specific goals instead of performing untargeted extraction. Therefore, we focus on \textbf{targeted training data extraction} in this paper, which requires recovering the suffix when given a prefix according to the training data. Compared with untargeted training data extraction, the task matters more because attackers can recover specific types of training data instead of any possible training data that might be harmless. What's more, it is easier to evaluate targeted training data extraction because we just need to compare the prediction with the ground truth suffix. However, for untargeted training data extraction, we need to search over the whole massive pre-training dataset (e.g., The Pile dataset \cite{gao2020pile}, which has 800GB text data) to check whether it contains the generated sample, which is very slow and costly.
% 看是否要提一个前缀只有一个可能的后缀的问题

The general process for targeted training data extraction can be divided into two steps: (1) generating one or more possible suffixes based on the given prefix, and (2) choosing a most likely suffix as the prediction result based on a confidence estimation method. We summarize two challenges of this task: (1) how to increase the generation likelihood of the ground truth suffix, and (2) how to estimate the confidence accurately so that the confidence score can be meaningfully interpreted as the probability that the output suffix is correct. To tackle these challenges, we propose a method named \textsc{Ethicist} for \textit{targeted training data \textit{\textbf{E}}xtraction \textit{\textbf{TH}}rough loss smoothed soft prompting and cal\textit{\textbf{I}}brated \textit{\textbf{C}}onf\textit{\textbf{I}}dence e\textit{\textbf{ST}}imation}.
For the first challenge, we propose loss smoothed soft prompting. It uses soft prompt to elicit memorization in the attacked model, and adds an additional loss besides the maximum likelihood estimation (MLE) loss to smooth the loss distribution of the suffix tokens. Through the loss smoothing, we hope to ensure that the probability of the ground truth token at each time step is not low, which makes it more likely to sample the ground truth suffix. With the two loss functions, we tune the prepended soft prompt tokens on an extracted training set which contains pairs of prefixes and ground truth suffixes. The existence of a training set is reasonable because large-scale pre-trained data generally contain public data (e.g., Common Crawl) \footnote{Similar setting is adopted in \citet{DBLP:journals/tacl/HisamotoPD20}.}. For the second challenge, we propose a calibrated confidence estimation method. We find that the model's perplexity cannot accurately represent the probability that the generated suffix is correct because the prediction probabilities for diversified prefixes are inherently different and incomparable. We thus normalize the confidence of the generated suffixes with a local estimation, which can mitigate the problems caused by intrinsic differences in the difficulties of distinct samples. We verify \textsc{Ethicist} on a recently proposed public benchmark containing 15,000 pairs of prefixes and suffixes derived from The Pile dataset \cite{gao2020pile}. 
% We conduct attacking experiments across various model sizes ranging from 125 million to 6 billion. 
Experiments show that \textsc{Ethicist} can significantly improve the extraction performance, which suggests that existing large language models are at significant risk of leaking training data. We also discuss and analyze several factors influencing the data extraction performance, including decoding strategy, model scale, prefix length, and suffix length.

Our contributions can be summarized as follows:
\begin{itemize}
    \item We propose loss smoothed soft prompting to reduce the difficulties of sampling the ground truth suffixes. 
    \item We propose a calibrated confidence estimation method that enables the confidence score to be meaningfully interpreted as the probability that the output suffix is correct.
    \item Experiments on a recently proposed benchmark demonstrate that \textsc{Ethicist} can consistently and significantly improve the data extraction performance across various model sizes. We further investigate several factors influencing the data extraction performance.
\end{itemize}

\section{Related Work}

\subsection{Training Data Extraction}
Existing works on training data extraction mainly focus on membership inference attack or untargeted training data extraction. For membership inference attack, adversaries need to judge whether a given example is contained in the training data of the attacked model. \citet{DBLP:conf/sp/ShokriSSS17, DBLP:conf/kdd/SongS19} train several shadow models that mimic the attacked models' behaviors to help train an auditing model that can predict whether an example is contained in the training dataset. \citet{DBLP:journals/tacl/HisamotoPD20} perform membership inference attacks on machine translation systems. They find it is harder to attack sequence generation models than classification models. \citet{DBLP:conf/ccs/SongR20} show that the encoded dense representations can leak information under membership inference attack. \citet{DBLP:journals/corr/abs-2203-03929} focuses on attacking masked language models that are pre-trained on possibly sensitive data (e.g., clinical notes). They introduce an additional reference masked language model besides the original attacked model and compute the ratio of the likelihood measured by the attacked model and the reference model, which is better than solely relying on the attacked model.

For untargeted training data extraction, adversaries first generate various samples using the attacked model and then predict whether they are contained in its training set. \citet{carlini2021extracting} extract hundreds of verbatim sequences from the popular pre-trained language model GPT-2 \cite{radford2019language}. And there is privacy information such as names, phone numbers, and email addresses in the extracted sequences. \citet{DBLP:conf/naacl/LehmanJPGW21} try to extract sensitive information from BERT \cite{devlin-etal-2019-bert} pre-trained on clinical notes. However, they are mostly unable to meaningfully expose Personal Health Information by simply using templates. Different from the existing works, we focus on targeted training data extraction that aims to recover the suffix when given a prefix, which is more security-critical and easier to evaluate. 

\subsection{Memorization}
We generally expect models can gain the generalization ability from the training process. However, recent works point out that models may unintentionally memorize the training data even without overfitting \cite{DBLP:journals/corr/abs-2205-10770, DBLP:journals/corr/abs-2202-07646, DBLP:conf/uss/Carlini0EKS19, DBLP:conf/ccs/BeguelinWTRPOKB20}. One possible method to mitigate this problem is to deduplicate training data \cite{DBLP:conf/icml/KandpalWR22}. However, \citet{DBLP:conf/uss/Carlini0EKS19} also show that it is possible to recover samples appearing only once in the training dataset. Surprisingly, \citet{DBLP:journals/corr/abs-2205-10770} find that there is a forgetting baseline during the pre-training of the casual language model (e.g., the model can memorize at least 40\% of the data that appear only once, even being trained on other data for many epochs afterward). These findings further emphasizes the difficulties of avoiding memorization and the potential threats of unintended memorization in large-scale pre-trained language models. Another line of work uses differential privacy to avoid the memorization problem \cite{DBLP:conf/ccs/AbadiCGMMT016, DBLP:conf/iclr/McMahanRT018, DBLP:conf/ccs/ShokriS15}, but the mechanism could reduce the accuracy \cite{DBLP:conf/uss/Jayaraman019, feldman2020neural, feldman2020does, DBLP:conf/kdd/SongS19}. Differential privacy also increases the training time, which can further influence the accuracy within the same budget. Therefore there is still no effective and practical way to avoid unintended memorization. Our work further verifies the existence of unintended memorization and makes it more necessary to develop practical defense methods.

\begin{figure*}[]
  \centering
  \includegraphics[width=\linewidth]{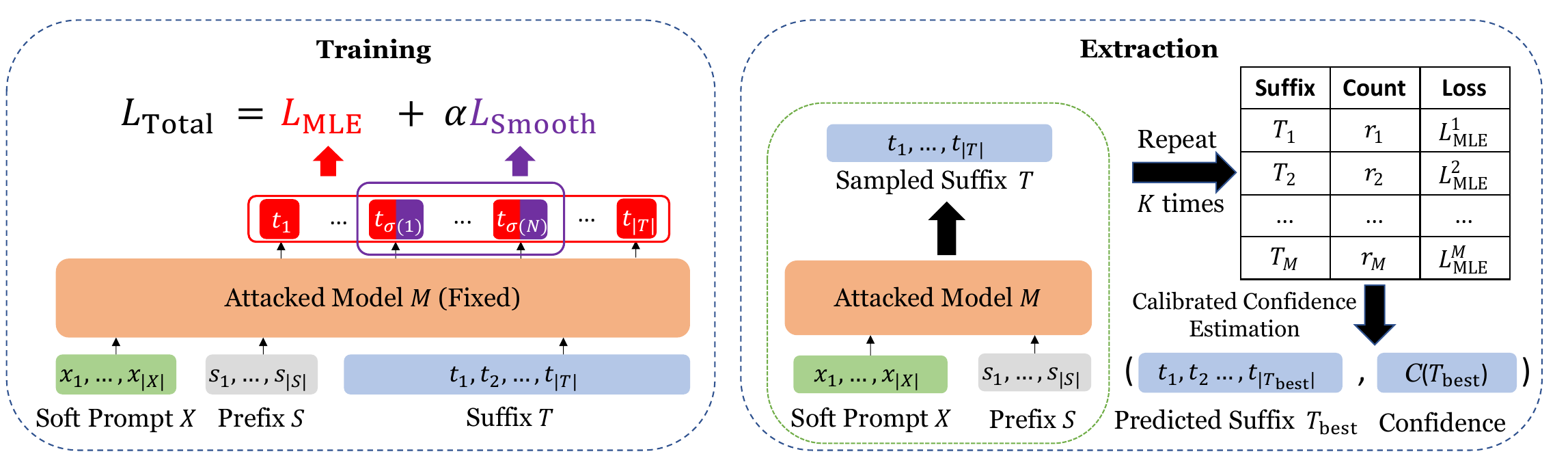}
  \caption{
    Method overview. During training, we fix the parameters of the attacked model $M$ and only tune the parameters of the soft prompt embeddings. Besides the MLE loss, we additionally design a smoothing loss to make the loss distribution of the suffix sequence more smooth. After tuning the soft prompt embeddings, we extract training data by repeatedly sampling K suffixes conditioned on one given prefix and using calibrated confidence estimation to select the final predicted suffix $T_{\text{best}}$ and provide its confidence $C(T_{\text{best}})$. }
  \label{fig:model}
\end{figure*}

\section{Methodology}

We formulate the targeted training data extraction task as follows: given a source prefix $S=(s_1,s_2,\cdots,s_{|S|})$ with $|S|$ tokens, the attacker should predict the target suffix $T=(t_1,t_2,\cdots,t_{|T|})$ with $|T|$ tokens and its confidence. The pair of the given prefix and the predicted suffix $(S,T)$ should be contained in the pre-training dataset  $D_{\text{pretrain}}=\{(S_i,T_i)\}$, which the attacked model $M$ is trained on. The prediction of the confidence score is necessary for picking out the most probable suffix when we don't know the ground truth suffix in realistic attack scenarios (i.e., we need to pick out most probable pairs of prefixes and extracted suffixes based on their confidence scores among all predictions). We assume the attacker can obtain some pairs of ground truth prefixes and suffixes $D_{\text{train}}=\{(S_i,T_i) |(S_i,T_i)\in D_{\text{pretrain}}, 1\leq i\leq |D_{\text{train}}|\}$  before attacking, which is reasonable because large-scale pre-trained data generally contain public data (e.g., Common Crawl). The attackers can utilize $D_{\text{train}}$ to train their attacking models and their goal is to predict suffixes for the prefixes in the test set $D_{\text{test}}=\{S_i | 1\leq i\leq |D_{\text{test}}|\}$. Note that the prefix $S_i$ in $D_{\text{test}}$ is included in $D_{\text{pretrain}}$ but is not a part of $D_{\text{train}}$.

\subsection{Method Overview}
An overview of \textsc{Ethicist} is shown in Figure \ref{fig:model}. We first tune the soft prompt embeddings during training to elicit memorization in the attacked model $M$ with the MLE loss and the additional smoothing loss. The smoothing loss aims to increase the probability of sampling the ground truth suffix. After prompt tuning, we repeatedly sample K suffixes using the attacked model $M$ conditioned on one given prefix and reorder them with our calibrated confidence estimation. Our calibrated confidence estimation can not only select the most possible suffix, but also provide a more accurate confidence score that represents how likely the predicted suffix is correct. Finally, the suffix with the highest confidence is selected as the final prediction.

\subsection{Prompt Tuning with Smoothing Loss}
We adopt prompt tuning to train the soft prompt tokens on $D$, which prepends $|X|$ soft tokens $X=(x_1,x_2,\cdots,x_{|X|})$ before the original input sequence. Then we feed the input to the attacked model $M$ to compute the MLE loss:
\begin{align}
    \mathcal{L}_{\text{MLE}}=\sum_{i=1}^{|T|}-\frac{1}{|T|}\text{log}P_M(t_i|X,S,t_{<i}).
\end{align}
Note that we only tune the parameters of the soft prompt tokens and the parameters of the attacked model $M$ are fixed. We use prompt tuning for two reasons: (1) we do not want to change the original parameters of the attacked model $M$ because the main goal is to elicit memorization in $M$, and (2) prompt tuning is helpful to improve the training efficiency when $M$ is very large, making \textsc{Ethicist} able to efficiently adapt to larger language models that generally memorize more training data.

The MLE loss aims to increase the total generation probability of the target suffix $T$. However, when using popular sampling methods such as top-k sampling \cite{fan2018hierarchical} and top-p (nucleus) sampling \cite{Holtzman2020The} to generate multiple candidate suffixes, we want to ensure the probability of the ground truth suffix token at \textbf{each time step} is not low. Suppose the total probability of the ground truth suffix is high while there is one token in the sequence with a low generation probability. In this case, it is still hard to generate the correct suffix using auto-regressive sampling methods. Therefore, we propose a smoothing loss to make the loss distribution of the suffix sequence more smooth. More specifically, we pick out the top-$N$ tokens with the highest loss values in the whole sequence $T$. Then we additionally optimize the generation probabilities for these $N$ tokens as follows:
\begin{align}
\label{smooth_loss}
\mathcal{L}_{\text{Smooth}}=\sum_{i=1}^{N}-\frac{1}{N}\text{log}P_M(t_{\sigma(i)}|X,S,t_{<\sigma(i)}),
\end{align}
where $t_{\sigma(i)}$ represents the token with the i-th highest loss in $T$. Note that $t_{\sigma(i)}$ is dynamically computed during training. The smoothing loss can also be seen as assigning higher weights to the tokens with higher loss values. Finally, we derive the overall loss function as follows:
\begin{align}
\label{total_loss}
    \mathcal{L}_{\text{Total}}=\mathcal{L}_{\text{MLE}}+\alpha\mathcal{L}_{\text{Smooth}},
\end{align}
where the coefficient $\alpha$ is a hyperparameter to control the strength of the smoothing loss.

\subsection{Calibrated Confidence Estimation}
After predicting the suffix, we also need to give a confidence score for the prediction, which can be meaningfully interpreted as the probability that the output suffix is correct. A naive method is to use the generation likelihood $P_T=\text{exp}(-|T|\mathcal{L}_{\text{MLE}})$ as the confidence score. This naive method is reasonable for picking out the most probable suffix $T_i$ from a collection of sampled suffixes $\{T_1,T_2,\cdots,T_M\}$ for one given prefix. However, it is unsuitable for comparing the confidence of different predicted suffixes corresponding to different prefixes. As the language model is essentially a statistical model, frequencies of tokens and n-grams in the prefixes can greatly influence the absolute generation likelihood of the suffixes. For example, consider two predicted suffixes $T_A$ and $T_B$ conditioned on two different prefixes $S_A$ and $S_B$, where $S_A$ and $T_A$ contain tokens and n-grams with much higher frequencies. The absolute generation likelihood of $T_A$ may be significantly higher than $T_B$, even if they are both ground truth suffixes. Therefore, to eliminate the intrinsic differences in scales of generation likelihood across different suffixes, we propose a novel calibrated confidence estimation method. To calibrate the confidence estimation, we have two considerations: (1) different generated suffixes conditioned on one given prefix should have comparable scales of generation likelihood, and (2) the memorized ground truth suffix is expected to be generated more frequently during multiple generations, which is also validated in Section \ref{sec:discussion}.

Suppose the sampled distinct suffixes are $\{T_1,T_2,\cdots,T_M\}$ for one given prefix, the repeated generation times for these suffixes are $\{r_1,r_2,\cdots,r_M\}$ (i.e., $r_i$ denotes how many times $T_i$ is generated among $K$ repeated sampling outputs), and the MLE loss values for these suffixes are $\{\mathcal{L}_{\text{MLE}}^1,\mathcal{L}_{\text{MLE}}^2,\cdots,\mathcal{L}_{\text{MLE}}^M\}$. Then we assign the calibrated confidence score to $T_i$ as:
\begin{align}
    C(T_i)=\frac{r_i\times\text{exp}(-|T_i|\mathcal{L}_{\text{MLE}}^i)}{\sum_{j=1}^M r_j\times\text{exp}(-|T_j|\mathcal{L}_{\text{MLE}}^j)}.
\end{align}
Through the proposed confidence estimation method, we obtain the confidence score of $T_i$ by comparing it with other sampled suffixes with comparable scales of generation likelihood. In this way, we avoid the scale problem brought by different prefixes and make it practical to compare the predicted suffixes conditioned on different prefixes. Moreover, we leverage the repetition time $r_i$ as a valuable signal since memorized suffix is expected to be generated more frequently. Finally, we select the suffix $T_{\text{best}}$ with the highest confidence score $C(T_{\text{best}})$ among $\{C(T_1),C(T_2),\cdots,C(T_M)\}$ as the predicted suffix and $C(T_{\text{best}})$ as its confidence estimation.

\section{Experiments}

\subsection{Benchmark}

We evaluate \textsc{Ethicist} on the LM-Extraction benchmark\footnote{\url{https://github.com/google-research/lm-extraction-benchmark/}}, which is designed for benchmarking targeted training data extraction attacks. It consists of a subset contained in The Pile dataset \cite{gao2020pile}. Both the prefix and the suffix are 50 tokens long.  All examples are well-specified, meaning that there is only one 50-token suffix in The Pile dataset given the 50-token prefix. What's more, these examples are all chosen to meet the property that there exists a prefix length (maybe longer than 50) that causes the model to generate the suffix string exactly, which implies that the extraction performance on this benchmark may be \textbf{higher} than that on randomly selected prefixes. We randomly split the dataset into training, validation and test sets. The detailed statistics of the LM-Extraction benchmark are shown in Table \ref{tab:data}.

\begin{table}[h]
    \centering
    \small
    \begin{tabular}{lccc}
    \toprule
    \textbf{Split} & \textbf{\# Examples} & \textbf{\# Prefix Len} & \textbf{\# Suffix Len} \\
    \midrule
    \textbf{Train} & 12,600 & 50 & 50 \\
    \textbf{Validation} & 1,400 & 50 & 50 \\
    \textbf{Test} & 1,000 & 50 & 50 \\
    \bottomrule
    \end{tabular}
    \caption{Statistics of the LM-Extraction benchmark.}
    \label{tab:data}
\end{table}

\subsection{Baselines}

 We compare \textsc{Ethicist} with the following baselines. All the compared baselines first sample K suffixes $\{T_1,T_2,\cdots,T_K\}$ conditioned on one given prefix $S$ and then pick out one suffix as the prediction.

\paragraph{Perplexity} It leverages the perplexity (PPL) measured by the attacked language model $M$ as the metric to sort the candidate suffixes and finally chooses the one with the lowest PPL as the predicted suffix $T$:
\begin{align*}
T=\arg\max_{T_i}C(T_i)=\arg\max_{T_i}\frac{1}{\text{PPL}_M(T_i|S)}
\end{align*}

\paragraph{Comparing (LM)} It takes another language model $M'$ and leverages the ratio of the perplexity measured by theses two language models as the metric \cite{carlini2021extracting}:
\begin{align*}
T=\arg\max_{T_i}C(T_i)=\arg\max_{T_i}\frac{\text{PPL}_{M'}(T_i|S)}{\text{PPL}_M(T_i|S)}
\end{align*}
The language model $M'$ could be a much smaller model trained on the same dataset with $M$ or trained on a different dataset. %The method supposes $\text{PPL}_M$ is relatively lower for the suffix $T_i$ memorized by $M$.  % Intuitively

\paragraph{Comparing (zlib)} Different from Comparing (LM), it uses the zlib \cite{gailly2004zlib} entropy of the text (i.e., the number of bits after compression with zlib) for comparison \cite{carlini2021extracting}:
\begin{align*}
T=\arg\max_{T_i}C(T_i)=\arg\max_{T_i}\frac{\text{len}(\text{zlib}(T_i))}{\text{PPL}_M(T_i|S)}
\end{align*}
% $$
% y = \arg\min_{1<=i<=K}\frac{PPL(s_i|\theta)}{\text{len}(zlib(s_i))}
% $$
% Text compressors can identify repeated patterns, which would also lead to a problematic low $\text{PPL}_M(T_i|S)$.

\paragraph{Comparing (lowercase)} It compares the perplexity of the original text and the lower-cased text measured by the same language model $M$ \cite{carlini2021extracting}:
\begin{align*}
T&=\arg\max_{T_i}C(T_i)\\
&=\arg\max_{T_i}\frac{\text{PPL}_{M}(\text{lowercased}(T_i)|S)}{\text{PPL}_M(T_i|S)}
\end{align*}
% $$
% y = \arg\min_{1<=i<=K}\frac{PPL(s_i|\theta)}{PPL(\text{lowercased}(s_i)|\alpha))}
% $$
Furthermore, we conduct ablation tests by removing the proposed components respectively to investigate the influence of each component.

% \paragraph{In-context Learning}

\subsection{Metrics}

We adopt the following automatic metrics for evaluation.

\paragraph{Recall} The metric computes the percentage of the  suffixes that are predicted verbatim over the whole test set. A higher recall score indicates better data extraction ability, which can also be understood as a higher attacking success rate.

\paragraph{Recall$_\text{Early stop}$} The metric first sorts the predictions according to their confidence scores and then evaluates the correctness of each prediction one by one. It finally computes the Recall score while making $x$ incorrect predictions. We set $x$ to 100 in our experiments following the LM-Extraction benchmark. A better confidence estimation method can give the correct predictions higher confidence scores and thus lead to a higher Recall$_\text{Early stop}$ score.

% Specifically, different from the standard Recall score that is calculated over the whole test set, Recall(Error) is calculated 

\subsection{Main Results}
Table \ref{tab:main_results} shows the automatic evaluation results with GPT-Neo 1.3B as the backbone. \textsc{Ethicist} achieves an impressive Recall score of 62.8\% and outperforms all the baselines by a large margin, indicating its better ability to extract training data from language models. Moreover, \textsc{Ethicist} has better confidence estimation performance after calibration as shown by a significantly higher Recall$_\text{Early stop}$ score. 
To further investigate the influence of each component, we run an ablation study. From the results shown in Table \ref{tab:main_results}, it can be seen that both the smoothing loss and the calibrated confidence estimation are important to enhance the ability to extract training data, and combining both of them achieves the best performance. Furthermore, we draw the following conclusions: (1) With prompt tuning and extra training data, we can better induce large-scale language models to generate their memorized training data and successfully achieves a 9.5\% performance improvement on Recall and a 12.4\% performance improvement on Recall$_\text{Early stop}$. (2) The proposed smoothing loss can further enhance the ability to extract training data, boosting the Recall score from 60.8\% to 62.3\%. (3) The calibrated confidence provides a 6.3\% improvement on Recall$_\text{Early stop}$ as expected, demonstrating the importance of calibrating confidence estimation for this task. (4) The smoothing loss is more effective in predicting exact suffixes while the calibrated confidence is more beneficial for identifying highly confident predictions, according to the significant drop in Recall without smoothing and the substantial decrease in Recall$_\text{Early stop}$ without calibration. (5) The calibrated confidence estimation is effective regardless of whether using prompt tuning. And it demonstrates greater advantages compared to the comparing (LM) baseline in recognizing predictions with higher confidence when using prompt tuning, indicated by increasing Recall$_\text{Early stop}$ (from 48.7 to 52.4).

\begin{table*}[t]
    \centering
    {
        \begin{tabular}{lcc}
        \toprule
        \textbf{Method} & \textbf{Recall} & \textbf{Recall$_\text{Early stop}$} \\
        \midrule
        % \multicolumn{3}{c}{\it{Fixed}}\\
        % \midrule
        \textbf{Perplexity} & 51.3 $_{\pm.0}$& 32.2 $_{\pm.0}$\\
        \textbf{Comparing (LM)} &51.9 $_{\pm.0}$ & 37.4 $_{\pm.0}$ \\
        \textbf{Comparing (zlib)} &49.7 $_{\pm.2}$ &25.6 $_{\pm.0}$ \\
        \textbf{Comparing (lowercase)} &51.5 $_{\pm.0}$ &32.5 $_{\pm.0}$ \\
        % \textbf{In-context Learning} & \\
        % \midrule
        % \multicolumn{3}{c}{\it{Fine-tuned}}\\
        \midrule
        \textbf{\textsc{Ethicist}}&\textbf{62.8 $_{\pm.5}$}&\textbf{53.8 $_{\pm.5}$} \\
        % \textbf{~~w/o prompt tuning} \\
        \textbf{~~w/o smooth}&61.2 $_{\pm.3}$ &52.4 $_{\pm.5}$ \\
        \textbf{~~w/o calibrated} &62.3 $_{\pm.6}$&47.5 $_{\pm1.3}$\\
        \textbf{~~w/o smooth \& calibrated} &60.8 $_{\pm.6}$&44.6 $_{\pm.8}$ \\
        \textbf{~~w/o smooth \& calibrated, comparing (LM)} &62.4 $_{\pm.7}$&48.7 $_{\pm1.2}$ \\
        \textbf{~~w/o prompt tuning} &50.9 $_{\pm.0}$&38.0 $_{\pm.0}$ \\
        \bottomrule
        \end{tabular}
    }
    \caption{Automatic evaluation results on the test set. The experiments are conducted on the GPT-Neo 1.3B model. We report the mean and the standard deviation over 3 random seeds. The best performance are highlighted in \textbf{bold}. \textbf{w/o smooth} means ablating the smoothed loss function in the training stage. \textbf{w/o calibrated} means ablating the calibrated confidence in the extraction stage. \textbf{w/o smooth \& calibrated} means prompt tuning only with the MLE loss and using the perplexity for confidence estimation. \textbf{~~w/o smooth \& calibrated, comparing (LM)} means prompt tuning only with the MLE loss and using the comparing (LM) method for confidence estimation. And \textbf{~~w/o prompt tuning} directly employs calibrated confidence estimation on the original model without prompt tuning.}
    \label{tab:main_results}
\end{table*}

\subsection{Analysis: Decoding Strategy}
% 采样方法分析
In our experiments, we use top-p sampling to sample multiple candidate suffixes conditioned on one given prefix. However, there are also other popular decoding methods, including greedy search, beam search, and top-k sampling. We thus compare these popular decoding methods in this section. Table \ref{tab:decode} shows the results. Not surprisingly, greedy search performs worst on both Recall and Recall$_\text{Early stop}$, which suggests some tokens in the ground truth suffix do not have the highest probability at the corresponding positions. Beam search outperforms top-p sampling on Recall, indicating that searching for the suffix with the lowest loss works well to find the ground truth suffix. However, beam search performs significantly worse than top-p sampling on Recall$_\text{Early stop}$, because it cannot use our calibrated confidence. Compared with beam search, top-p sampling can generate multiple candidates, which could substantially increase the accuracy of confidence estimation with our proposed calibrated confidence. Moreover, the top-k sampling performs worse than top-p sampling on Recall$_\text{Early stop}$, which may be because top-k sampling is easier to sample low-probability tokens and thus reduce the confidence of the ground truth suffixes. We finally select top-p sampling as our decoding method due to its balance on Recall and Recall$_\text{Early stop}$.

\begin{table}[!t]
    \centering
    {
        \begin{tabular}{lcc}
        \toprule
        \textbf{Strategy} & \textbf{Recall} & \textbf{Recall$_\text{Early stop}$}  \\
        \midrule
        \textbf{Greedy} & 58.7 $_{\pm.6}$ & 47.1 $_{\pm1.1}$ \\
        \textbf{Beam Search} & 64.5 $_{\pm.9}$ & 47.9 $_{\pm1.0}$\\
        \textbf{Top-k} & 62.7 $_{\pm.6}$ & 50.8 $_{\pm.6}$ \\
        \textbf{Top-p} & 62.8 $_{\pm.5}$ & 53.8 $_{\pm.5}$ \\
        \bottomrule
        \end{tabular}
    }
    \caption{Effect of the decoding strategy on \textsc{Ethicist}. Note that our proposed calibrated confidence is unused when decoding with deterministic methods, including greedy search and beam search. We show the mean and the standard deviation over 3 random seeds for all decoding strategies.} 
    % The beam size is set to 10. The k for top-k sampling is set to 10. The p for top-p sampling is set to 0.7. The temperature is set to 0.8 for both top-k and top-p sampling. 
    \label{tab:decode}
\end{table}

\subsection{Analysis: Model Scale}
% *预训练模型大小的影响*

\begin{figure}[!t]
  \centering
  \includegraphics[width=\linewidth]{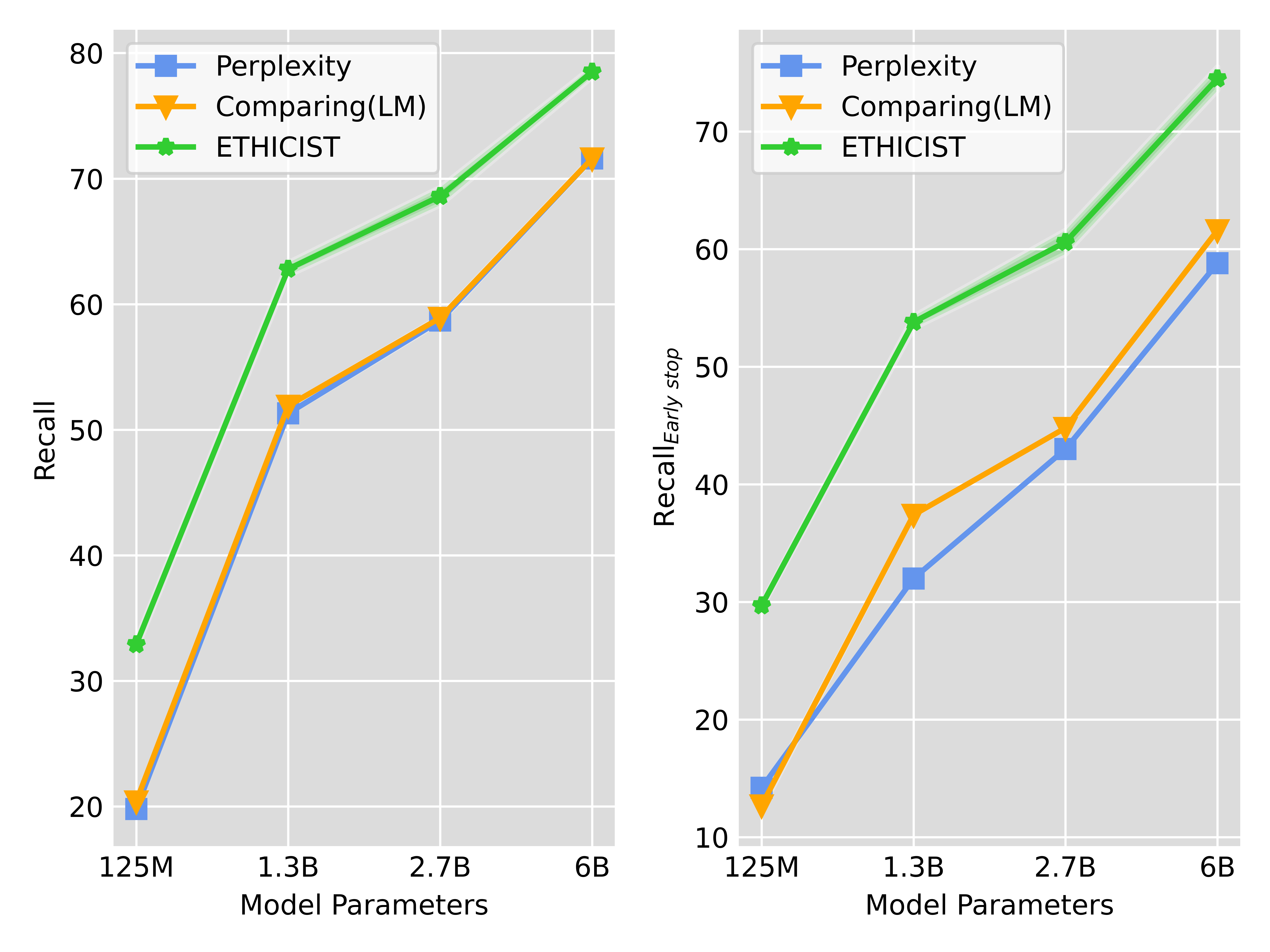}
  \caption{
    Effect of the model scale. We show the mean and the standard deviation over 3 random seeds for all methods.
  }
  \label{fig:model_scale}
\end{figure}

Previous works on scaling laws find that larger language models can memorize more training data \cite{carlini2022quantifying, tirumala2022memorization}. Therefore, we are interested in how targeted data extraction performance varies across different model scales. 
% Moreover, we focus on precisely quantifying how much language models memorize rather than just reporting a loose memorize-versus-scale relationship (e.g., a positive correlation which is known as a prior). 
Figure \ref{fig:model_scale} shows the results. We can see that the targeted training data extraction performance continuously increases as the model scale increases from 125 million to 6 billion. \textsc{Ethicist} shows impressive results as it consistently and significantly outperforms baselines across different model scales. Thanks to prompt tuning, \textsc{Ethicist} is efficient in terms of computation time and particularly memory consumption. Therefore, \textsc{Ethicist} can also be adapted to larger language models for efficient targeted training data extraction.

% \begin{table}[t]
%     \centering
%     {
%         \begin{tabular}{lcc}
%         \toprule
%         \textbf{Method} & \textbf{Recall} & \textbf{Recall$_\text{Early stop}$}\\
%         \midrule
%         \multicolumn{3}{c}{\it{GPT-Neo 125M}}\\
%         \midrule
%         \textbf{Perpelxity} & 19.8 $_{\pm.0}$ & 14.2 $_{\pm.0}$ \\
%         \textbf{Ours} & 32.9 $_{\pm.4}$ & 29.7 $_{\pm.2}$ \\
%         \midrule
%         \multicolumn{3}{c}{\it{GPT-Neo 1.3B}}\\
%         \midrule
%         \textbf{Perplexity} & 51.3 $_{\pm.0}$ & 32.2 $_{\pm.0}$ \\
%         \textbf{Ours} & 62.8 $_{\pm.5}$ & 53.8 $_{\pm.5}$ \\
%         \midrule
%         \multicolumn{3}{c}{\it{GPT-Neo 2.7B}}\\
%         \midrule
%         \textbf{Perplexity} &  58.7 $_{\pm.1}$ & 43.0 $_{\pm.0}$\\
%         \textbf{Ours} & 68.6 $_{\pm.7}$ & 60.6 $_{\pm1.0}$\\
%         \midrule
%         \multicolumn{3}{c}{\it{GPT-J 6B}}\\
%         \midrule
%         \textbf{Perplexity} & 71.6 $_{\pm.0}$ & 58.8 $_{\pm.0}$\\
%         \textbf{Ours} & 78.5 $_{\pm.4}$ & 74.5 $_{\pm.1}$\\
%         \bottomrule
%         \end{tabular}
%     }
%     \caption{Model Scale}
%     \label{tab:model scale}
% \end{table}

\begin{figure}[!t]
  \centering
  \includegraphics[width=\linewidth]{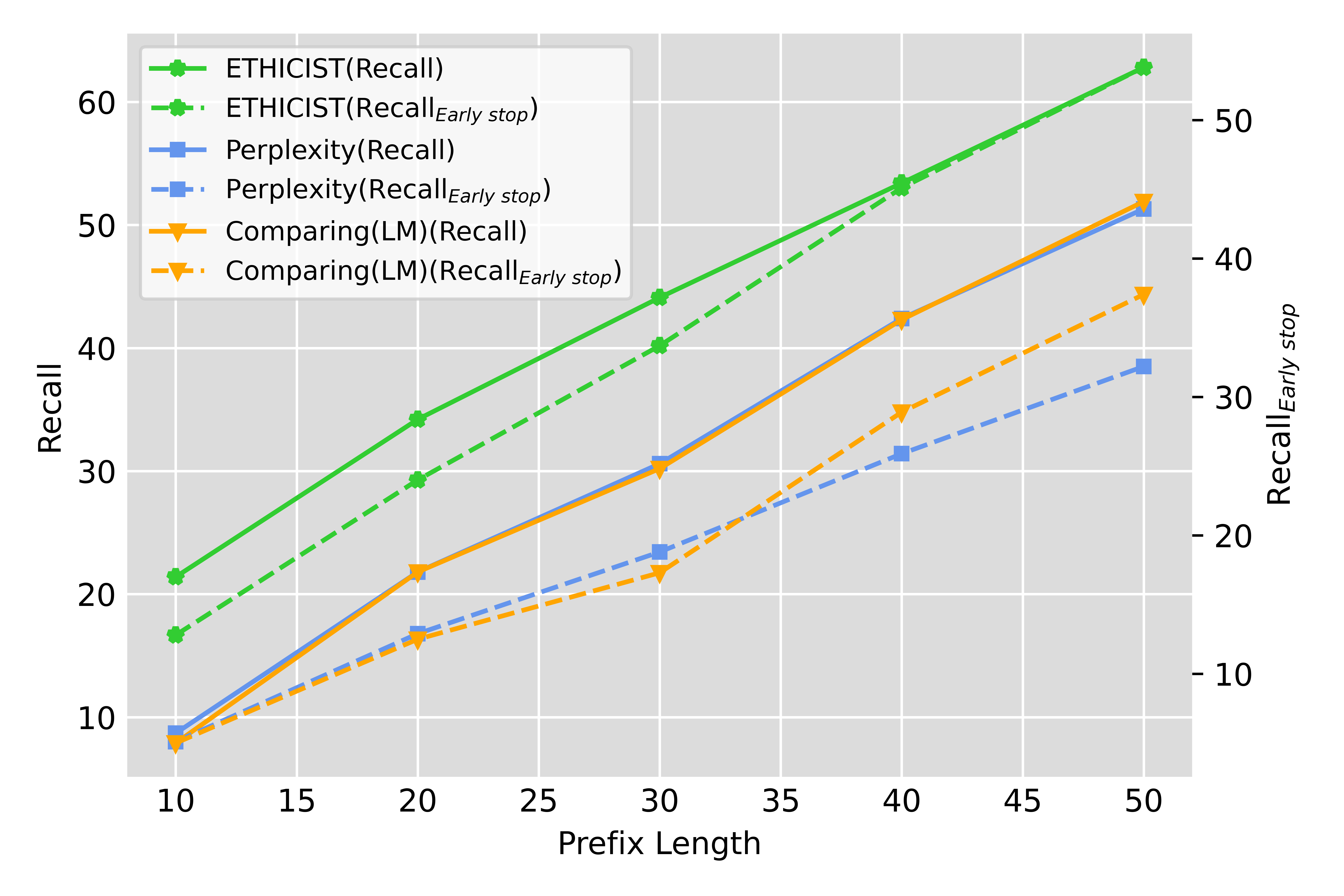}
  \caption{
    Effect of the given prefix length. We show the \textbf{Recall} and the \textbf{Recall$_\text{Early stop}$} metrics for three methods when the given prefix length increases from 10 to 50. 
  }
  \label{fig:prefix_len}
\end{figure}

\begin{figure}[!t]
  \centering
  \includegraphics[width=\linewidth]{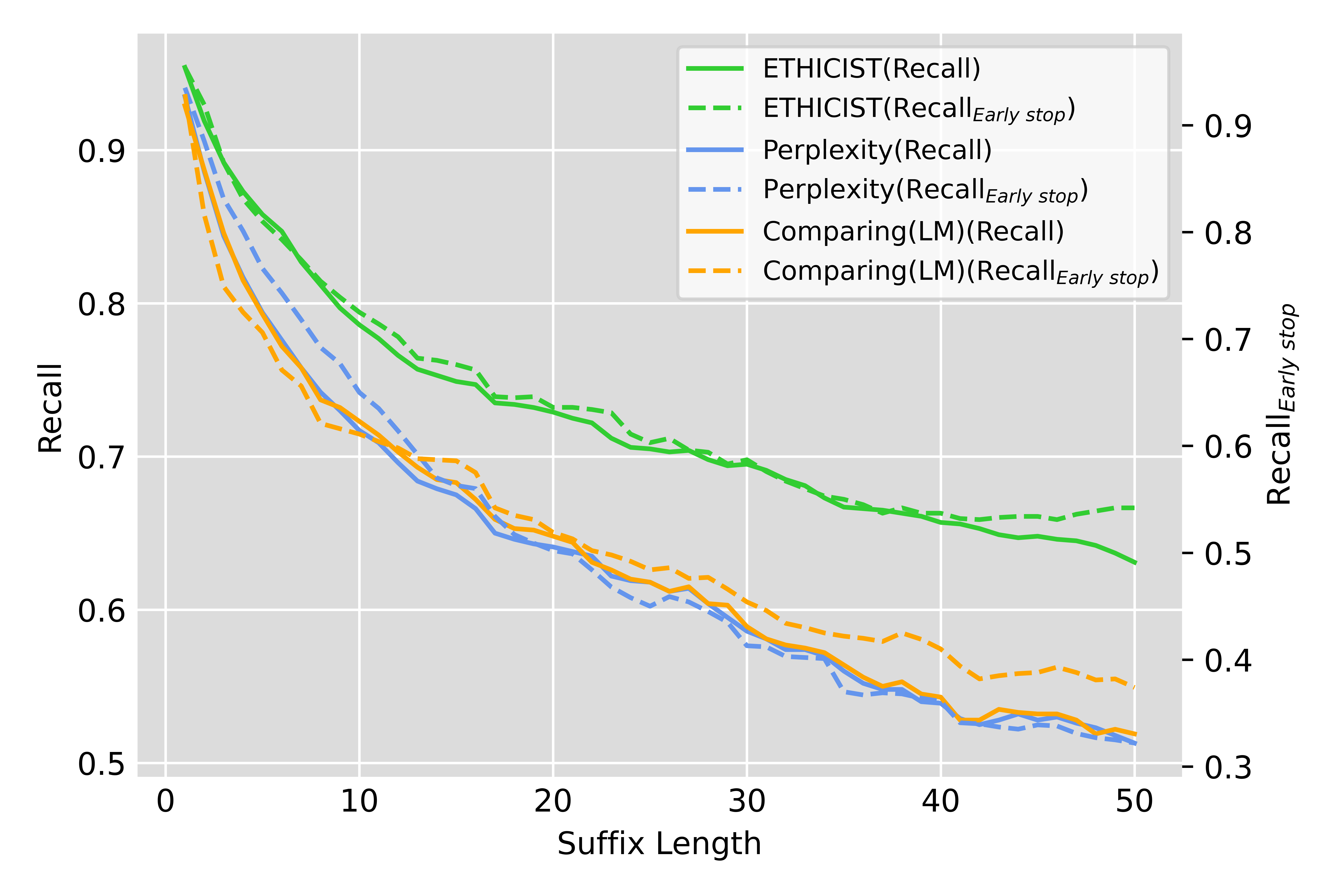}
  \caption{
    Effect of the predicted suffix length. We show the \textbf{Recall} and the \textbf{Recall$_\text{Early stop}$} metrics when the predicted suffix length increases from 1 to 50. 
  }
  \label{fig:suffix_len}
\end{figure}

\subsection{Analysis: Prefix Length and Suffix Length}
% 生成的后缀和给定的前缀的长度的影响
All prefixes and suffixes in the LM-Extraction benchmark are 50 tokens long, making it an interesting question how the length of prefixes and suffixes would affect the extraction performance. 

We show the effect of the given prefix length in Figure \ref{fig:prefix_len}. We can observe that the extraction performance grows approximately linearly with the prefix length for all evaluated methods, and \textsc{Ethicist} performs best for all prefix lengths. Although all methods have similar growth speed on Recall, \textsc{Ethicist} has the highest growth speed on Recall$_\text{Early stop}$. It is also interesting that \textit{Comparing (LM)} only outperforms \textit{Perplexity} when given prefixes that are long enough.

We show the effect of the predicted suffix length in Figure \ref{fig:suffix_len}. For all three methods, the extraction performance decreases when the suffix length increases. Different from the approximately linear relationship between the prefix length and the extraction performance, the performance degradation tends to become progressively slower as the suffix length increases. This suggests that the model can still memorize a considerable proportion of suffixes (rather than quickly decreasing to zero) even if the predicted suffix length continues to increase. What's more, we observe that \textsc{Ethicist} has a significantly slower speed of performance degradation compared with the two baselines, which suggests \textsc{Ethicist} is effective for eliciting deeper memorization of longer suffixes of the attacked model.

\subsection{Analysis: Sampling Time}

Due to space limitations, we put the analysis of sampling time in Appendix \ref{sec:sample_time}.

% \subsection{Analysis: Memorized Data}
% easy case: 被记忆数据的特点（例如生成重复次数）
% hard case

% \subsection{Case Study}
% case展示，比如隐私信息的数据

\begin{table}[!t]
    \centering
    {
        \begin{tabular}{lcc}
        \toprule
        \textbf{Feature} & \textbf{Correct} & \textbf{Wrong}  \\
        \midrule
        \textbf{Recall@1} & 0.63 & 0.37 \\
        \textbf{Recall@3} & 0.68 & 0.32 \\
        \textbf{Recall@100} & 0.69 & 0.31 \\
        \textbf{Average Repeat Time} & 85.38 & 29.66 \\
        \textbf{Average Confidence} & 0.95 & 0.67 \\
        \bottomrule
        \end{tabular}
    }
    \caption{Statistical features of correct predictions and wrong predictions. Recall@K measures whether the top-K suffixes sorted by estimated confidence contain the ground truth suffix. Average repeat time represents the number of times that the prediction result is generated repeatedly out of 100 generations.}
    \label{tab:statistic}
\end{table}

\begin{figure}[]
  \centering
  \includegraphics[width=\linewidth]{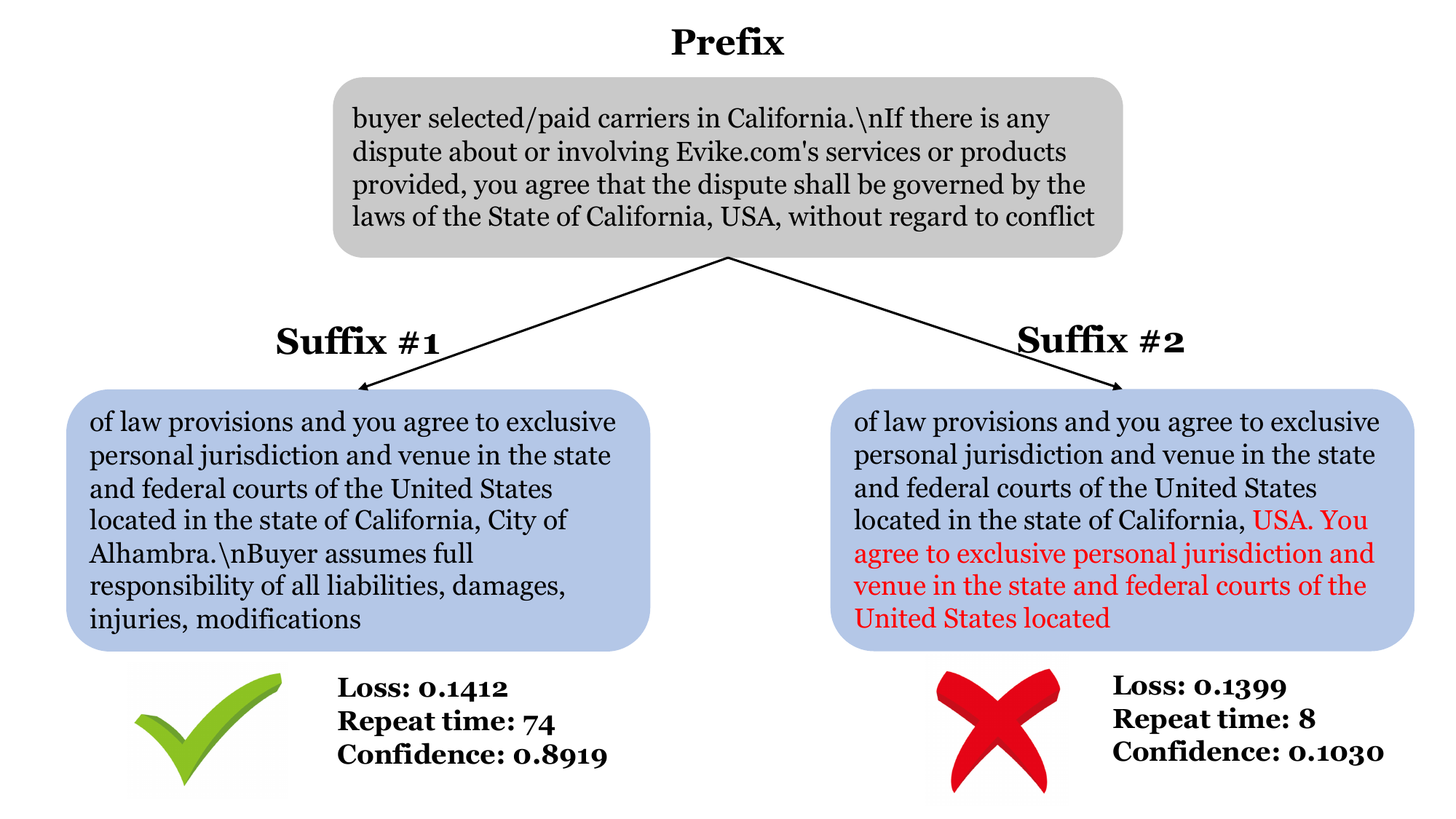}
  \caption{
    Given the prefix, we show the top-2 predicted suffixes by \textsc{Ethicist}. Although the first prediction has higher loss, it is repeated for 74 times and we correctly select it as the final predicted suffix using our calibrated confidence estimation. We highlight the wrong predicted text in \textcolor{red}{red}.
  }
  \label{fig:loss_case}
\end{figure}

\begin{figure}[!h]
  \centering
  \includegraphics[width=\linewidth]{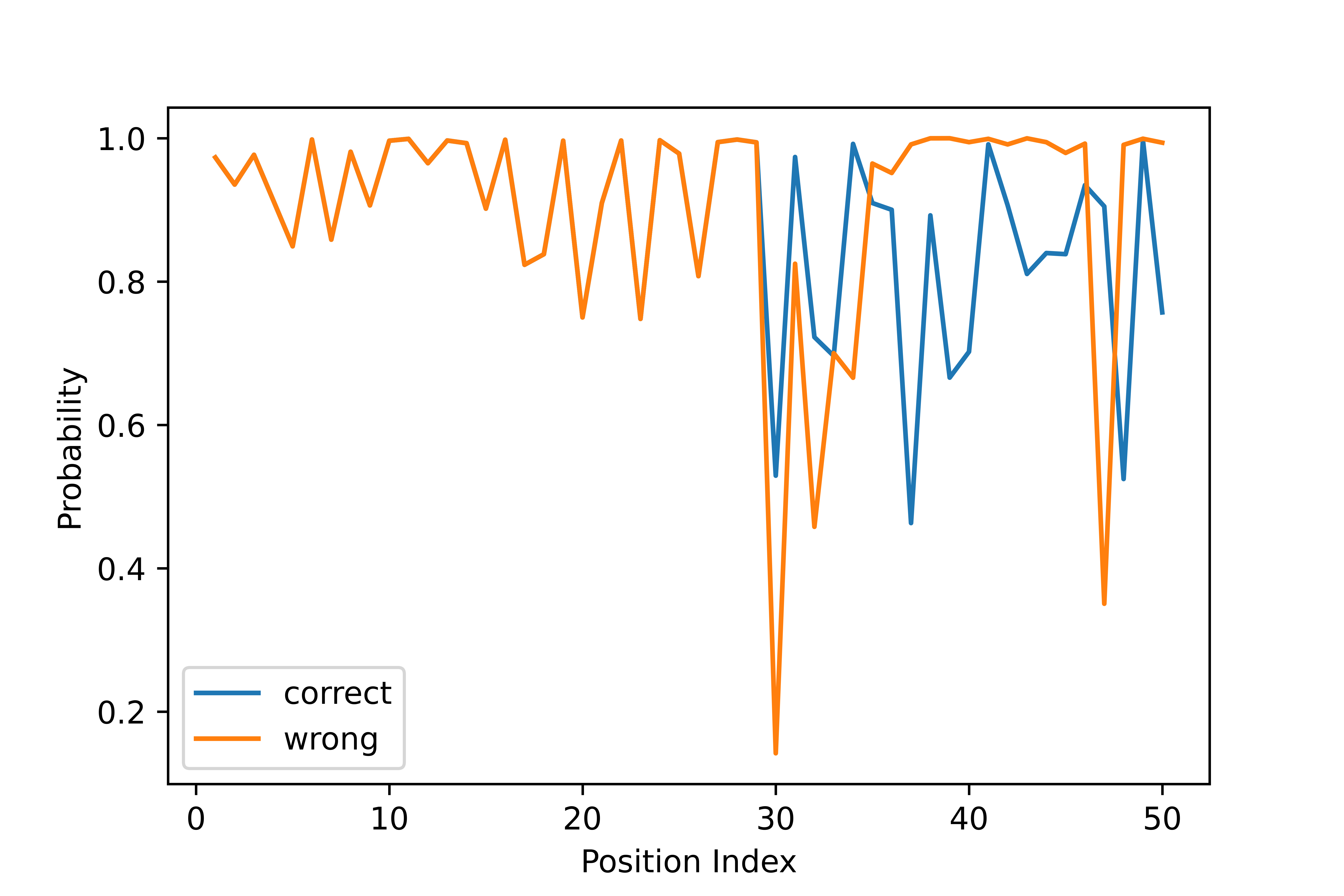}
  \caption{
    We show the generation probability of each token during the sampling process for both correct and wrong suffixes.
  }
  \label{fig:loss_plot}
\end{figure}

\section{Discussion} \label{sec:discussion}
We further show some statistical features in Table \ref{tab:statistic}. We can see that the memorized suffixes can be sampled significantly more frequently with a high average repeat time of 85.38, validating that the repeat time is a valuable signal for confidence estimation. What's more, the memorized suffixes have significantly higher confidence. One interesting phenomenon we observe is that if the ground truth suffix can be generated, it mostly has the top 3 highest confidence (Recall@3 $\approx$ Recall@100). We also find that for more than 30\% of the prefixes, the model cannot generate the correct prefix even given 100 chances. Therefore, an important future direction is to design better methods to elicit memorization in the attacked model. Considering the non-negligible gap between Recall@1 and Recall@100 (0.63 vs. 0.69), another important future direction is to design better confidence estimation methods (maybe trainable), which can pick out the ground truth suffix among the collection of candidate suffixes for one prefix.

We show a case in Figure \ref{fig:loss_case}. Although the first predicted suffix has higher loss than the second predicted suffix, it is sampled far more times than the latter. Therefore, we assign higher confidence to the first suffix using our calibrated confidence estimation method. We further show the probability of generating each token during the sampling process in Figure \ref{fig:loss_plot}. We can observe that although the correct prediction has higher loss as a whole, it keeps a high sampling probability across the generation process. The minimum probability of generating one token in the correct suffix is about 0.45, which is significantly higher than 0.1 for the wrong suffix. Therefore it is easier to generate the correct suffix, which leads to a higher confidence score. This is also in line with our motivation for designing the extra smoothing loss, which can increase the probability of sampling the correct suffix.

\section{Conclusion}
In this work, we propose \textsc{Ethicist}, an effective method for targeted training data extraction attack. \textsc{Ethicist} uses soft prompt to elicit memorization in the attacked model. To ensure the probability of the ground truth suffix token at each time step is not low, we propose a smoothing loss besides the standard MLE loss.
We also propose a calibrated confidence estimation method to calibrate the scale of confidence across different samples.
% We also point out that the original perplexity given by the model is not comparable across different examples and cannot accurately represent the probability that the generated suffix is correct. Therefore, \textsc{Ethicist} uses a calibrated confidence estimation method which not only calibrates the scale of confidence across different samples, but also leverages the repeatedly generation time as a valuable signal.
Experiments on the LM-Extraction benchmark demonstrate that \textsc{Ethicist} significantly improves the extraction performance. We further conduct extensive experiments to investigate several critical factors influencing the extraction performance, including decoding strategy, model scale, prefix length, and suffix length.
% We find that top-p sampling is the most balanced sampling method for this task. We also find that the longer the prefix and the shorter the suffix, the better the attack performance. Similar to the findings of previous works, we also observe that bigger models can memorize more training data, which suggests the seriousness of the information leakage problem. 
We hope our work can promote future researches on better attack methods and practical defense methods for the training data extraction problem.

\section*{Acknowledgement}
This work was supported by the NSFC projects (Key project with No. 61936010 ). This work was also supported by the Guoqiang Institute of Tsinghua University, with Grant No. 2020GQG0005.

\section*{Limitations}
Although we conduct experiments across various model scales ranging from 125M to 6B, there are still larger language models we don't test either because their training data is not publicly released or because we have limited resources. 

Moreover, the examples in the LM-Extraction benchmark are all chosen to meet the property that there exists a prefix length (maybe longer than 50) that causes the model to generate the suffix string exactly, which makes the extraction performance on this benchmark higher than that on randomly selected prefixes. 

% Our experiments are conducted on an English dataset, but our method is not limited to specific languages. 

\section*{Ethics Statement}
\textsc{Ethicist} is a powerful method to elicit memorization in the large pre-trained language models, which makes it a useful tool to expose the privacy risk of large language models. However, it also has a risk to be abused by attackers to extract privacy information from pre-trained language models. Thus large language models should be carefully examined before being made publicly available. What's more, it is necessary to develop defense methods against the training data extraction attacks without sacrificing the language modeling ability. 

The LM-Extraction benchmark is derived from the Pile dataset, and thus covers many domains including books, code, emails, etc. This suggests the effectiveness of targeted training data extraction across different domains.

\bibliography{anthology,custom}
\bibliographystyle{acl_natbib}

\appendix

\section{Implementation Details}

As the benchmark is derived from The Pile \cite{gao2020pile} dataset, we conduct experiments only on the models that are pre-trained on The Pile dataset. They are GPT-Neo 125M, GPT-Neo 1.3B, GPT-Neo 2.7B, and GPT-J 6B \cite{gpt-neo, gpt-j}. We set the prompt length to 100, the batch size to 32, the learning rate of AdamW optimizer to 1e-3, the warmup step to 500, the learning rate decay strategy to linear, $N$ in Equation \ref{smooth_loss} to 5, $\alpha$ in Equation \ref{total_loss} to 0.7, and the maximum training epoch to 20 with an early stopping mechanism. In our main experiments, we generate the suffix using top-p sampling \cite{Holtzman2020The} with $p=0.7$ and $\text{temperature}=0.8$. For other decoding methods, we set beam size to 10 for beam search, and k to 10 for top-k sampling ($\text{temperature}=0.8$). Our code is based on Huggingface Transformers \cite{wolf-etal-2020-transformers}.

\section{Computing Infrastructure}
All experiments are carried out on a single Tesla V100 GPU with 32GB memory. Each experiment can be completed in less than 20 hours.

\begin{figure}[!h]
  \centering
  \includegraphics[width=\linewidth]{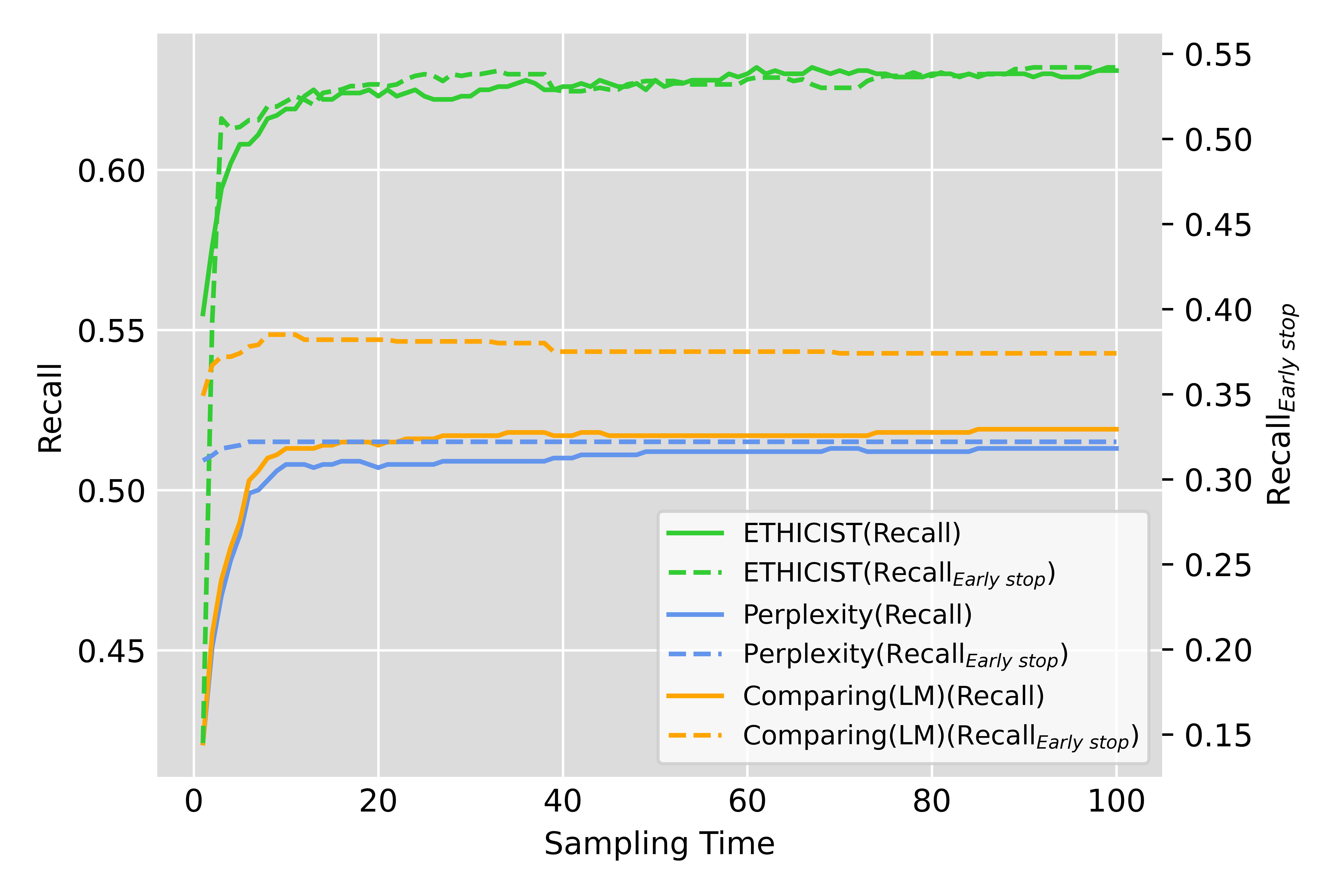}
  \caption{
    Effect of the sampling time. We show the \textbf{Recall} and the \textbf{Recall$_\text{Early stop}$} metrics for three methods when the sampling time increases from 1 to 100. 
  }
  \label{fig:sample_time}
\end{figure}

\section{Effect of Sampling Time}
\label{sec:sample_time}
In our main experiments, we sample 100 candidate suffixes for one given prefix. We show the effect of sampling time in Figure \ref{fig:sample_time}. We can see that all methods' performances increase quickly when the sampling time increases from 1 to 10. However, \textsc{Ethicist}'s performance can still improve slowly when the sampling time increases from 10 to 100, which we attribute to the consideration of repeat time in our calibrated confidence estimation. What's more, although we report the result for sampling 100 times in our main experiments, we can see that \textsc{Ethicist} can achieve satisfying performance when sampling only 10 times, which suggests the efficiency of \textsc{Ethicist}.

\end{document}